\documentclass{article}
 \usepackage[]{lineno}
\usepackage{setspace}
\usepackage{PRIMEarxiv}
\usepackage{subfigure}
\usepackage[utf8]{inputenc} % allow utf-8 input
\usepackage[T1]{fontenc}    % use 8-bit T1 fonts
\usepackage{hyperref}       % hyperlinks
\usepackage{url}            % simple URL typesetting
\usepackage{booktabs}       % professional-quality tables
\usepackage{amsfonts}       % blackboard math symbols
\usepackage{nicefrac}       % compact symbols for 1/2, etc.
\usepackage{microtype}      % microtypography
\usepackage{lipsum}
\usepackage{fancyhdr}       % header
\usepackage{graphicx}       % graphics
\graphicspath{{media/}}     % organize your images and other figures under media/ folder
\usepackage{float}
\usepackage{graphicx}
\usepackage{longtable}
\usepackage{multirow}
\usepackage{ragged2e}
\usepackage{makecell}
\usepackage{enumitem}
\usepackage{apacite}%%
\usepackage{natbib}%%
\title{Real-time Strawberry Detection Based on Improved YOLOv5s Architecture for Robotic Harvesting in open-field environment
}

\author{
  Zixuan He, Salik Ram Khanal \\
 Center for Precision and Automated Agricultural Systems \\
  Department of Biological Systems Engineering\\
  Washington State University \\
 Prosser, WA 99350\\
 \\
 \And
   Xin Zhang \\
  Department of Agricultural and Biological Engineering\\
  Mississippi State University \\
  Mississippi State, MS 39762\\
  \\
  \And
   Manoj Karkee, Qin Zhang \\
 Center for Precision and Automated Agricultural Systems \\
  Department of Biological Systems Engineering\\
  Washington State University \\
 Prosser, WA 99350\\
   \texttt{\{Corresponding author:Manoj Karkee\} \url{manoj.karkee@wsu.edu}}
   \\
}

\begin{document}
\maketitle

\begin{abstract}
Strawberry detection is one of the crucial steps during vision-guided robotic harvesting, which provides the precise location, pose, and maturity of the fruit to the manipulation system for picking. False detection and/or localization of strawberries with multiple maturity levels result in poor performance of robotic picking and/or causes fruit injury or unnecessary picking on fruits in wrong maturities in immature or nearly mature stages during robotic harvesting. This study proposed a YOLOv5-based custom object detection model to detect strawberries in an outdoor environment. The original architecture of the YOLOv5s was modified by replacing the C3 module with the C2f module in the backbone network, which provided a better feature gradient flow. Secondly, the Spatial Pyramid Pooling Fast in the final layer of the backbone network of YOLOv5s was combined with Cross Stage Partial Net to improve the generalization ability over the strawberry dataset in this study. The proposed architecture was named YOLOv5s-Straw.  The RGB images dataset of the strawberry canopy with three maturity classes (immature, nearly mature, and mature) was collected in open-field environment and augmented through a series of operations including brightness reduction, brightness increase, and noise adding. To verify the superiority of the proposed method for strawberry detection in open-field environment, four competitive detection models (YOLOv3-tiny, YOLOv5s, YOLOv5s-C2f, and YOLOv8s) were trained, and tested under the same computational environment and compared with YOLOv5s-Straw. The results showed that the highest mean average precision (mAP) of 80.3$\%$ was achieved using the proposed architecture whereas the same was achieved with YOLOv3-tiny, YOLOv5s, YOLOv5s-C2f, and YOLOv8s were 73.4\%, 77.8\%, 79.8\%, 79.3\%, respectively. Specifically, the average precision of YOLOv5s-Straw was 82.1\% in the immature class, 73.5\% in the nearly mature class, and 86.6\% in the mature class, which were 2.3\% and 3.7\% higher than that of the latest YOLOv8s,  respectively for mature and immature classes. The proposed model was lighter and faster with 8.6$\times$10$^{6}$ network parameters and an inference speed of 18ms per image compared with YOLOv8s model which had an inference speed of 21.0ms and parameters of 11.1$\times$10$^{6}$. The proposed model showed strong potential for providing accurate positions of mature strawberries for developing an effective strawberry harvesting system in an outdoor environment.
\end{abstract}

% keywords can be removed
\keywords{ artificial intelligence \and strawberry detection \and YOLOv5 \and fruit maturity  \and robotic harvesting}

\section{Introduction}
Traditionally, strawberries are harvested manually, but this is becoming more challenging with increased labor costs and labor shortages from an aging workforce in the United States and around the world \citep{delbridge2021robotic}. The shortage of field workers in agriculture was reported in recent years, especially during the COVID-19 pandemic including in fruit crop farming \citep{charlton2021potential}. To address this challenge, automated or robotic strawberry harvesting methods are being investigated as a promising alternative to manual harvesting. As a fundamental component of automated/robotic strawberry harvesting, a machine vision system needs to be designed to work effectively on detecting the target fruit, including location, maturity, and quality, under natural lighting conditions often characterized by uncertain and variable light intensities over space and time \citep{he2022detecting, yan2021real}.

Most of the studies on strawberry detection used image processing methods, such as segmentation \citep{feng2008fruit}, edge detection \citep{liming2010automated}, and feature extraction in different color spaces \citep{shen2009image}. The detection using these digital image processing methods required stable lighting conditions, therefore, it was challenging to achieve high accuracy in a real-time outdoor environment with these techniques. Additionally, these methods applied manually-designed feature detection techniques, which did not perform well when facing various morphological changes, e.g., multiple maturities and shapes of strawberries, in the field environment. Adopting traditional image processing methods is, therefore, insufficient to meet the requirements of a harvesting robot.

With the advancement of convolutional neural networks (CNN) and computational hardware including powerful graphical processing units, scientists have been working on alternative approaches for developing robust machine vision systems for agricultural applications. CNN makes use of convolution operations to predict and classify the target object in an image. The major advantage of CNN-based vision systems is the ability to fine-tune the pre-trained network parameters to adapt to custom training samples. CNN offers robustness in the generalization of object identification with different shapes, sizes, colors, and textures, and backgrounds with complex structures. A number of CNN-based approaches were studied for object detection, segmentation, and localization in various crop canopies. \cite{zhou2017classification} proposed a classification and grading approach for tomatoes using a Fast-RCNN architecture. A similar approach for fruit detection was proposed by \cite{hussain2018automatic} to detect various types of fruit in a commercial farm. Based on the region detection architecture, a Fast-RCNN model with Zeiler and Fergus Net was applied for detecting apples, which achieved a recognition rate of 92.3\% \citep{fu2018kiwifruit}. Similarly, detection, used by \cite{chen2019strawberry}, of the whole strawberry field based on Faster-RCNN was used for yield prediction, which had faster interference speed than Fast-RCNN to improve harvesting efficiency. Although these techniques reported relatively high accuracy and robustness in detecting fruit in variable lighting conditions, it was still limited to meet the needs of real-time detection due to the heavy size of these CNN-based models with a great number of trainable parameters.

The You-Only-Look-Once (YOLO) models \citep{redmon2016you} detect objects and provide rectangular bounding boxes around those objects with a class probability using a single feed-forward network, making deep learning models computationally much more efficient and faster. Researchers are continuously working on YOLO object detection techniques and reporting improved performances over various datasets. YOLOv3 \citep{redmon2018yolov3} is one of the preferred networks for object detection in various applications because of its better computational performance and detection accuracy, allowing machine vision systems to achieve real-time object (e.g., fruit) detection. Researchers have achieved promising results in fruit detection by adopting the YOLOv3 structure \citep{yu2020real,liu2020yolo}. With an improvement over YOLOv3 by introducing spatial pyramid pooling and path aggregation network, it was demonstrated that YOLOv4 achieved an Average Precision (AP) of 43.5\% in the MS COCO dataset and 65 fps processing time on a Tesla V100 GPU \citep{bochkovskiy2020yolov4}. Later, YOLOv5 was introduced with a similar structure to YOLOv4 but containing additional features of mosaic for data augmentation and auto-learning bounding boxes anchors, which could improve the detection performance with lower learning loss \citep{Jocher_YOLOv5_by_Ultralytics_2020}. At present, the YOLOv8 is the latest release of the YOLO family with new C2f modules and the anchor-free function, which improves the detection performance but the reported interference speed is slower than that of YOLOv5 under the same computation environment due to increased parameters in YOLOv8\citep{Jocher_YOLO_by_Ultralytics_2023}. The primary characteristic of the YOLO family is the rapid inference speed while exhibiting satisfactory performance in detecting small objects in complex environments, which is suitable for the strawberry detection task in robotic harvesting.

In robotic strawberry harvesting, strawberries with their soft surface can be easily damaged or bruised during the picking process from the impact caused by mechanical/robotic parts \citep{yu2020real}, particularly when the strawberries are detected inaccurately (e.g., shifted center location). Handling various obstacles is another challenge in the strawberry canopies. In addition, inconsistent maturities of strawberries in the canopy do not allow a robotic harvester to pick them all at the same time. Therefore, strawberry detection systems for robotic picking need to be precise and accurate and should be able to recognize multiple maturity levels of strawberries in the field environment, and varying light conditions. Recent models in the area of strawberry detection were developed using CNN-based approaches. Many CNN models, such as YOLOv3 \citep{yu2020real}, YOLOv4 \citep{zhang2022real}, YOLOv5 \citep{he2022detecting,fan}, Faster-RCNN \citep{lamb2018strawberry}, and Mask-RCNN \citep{ge2019fruit,yu2019fruit}, were successfully adopted in strawberry detection for robotic harvesting but most of them were experimented in greenhouses and polytunnels environment. The critical challenges in the field environment (e.g., shadows, multiple maturity detection, multiple light conditions) have not yet been fully studied for strawberry detection with multiple maturities under outdoor environments. On the other hand, present technologies based on deep learning have sufficient accuracy for strawberry detection but have still been limited in flexibility due to high computation cost (heavy models with a large number of trainable parameters) which would further impact the efficiency of harvesting robots. It is, therefore, necessary to develop a lightweight strawberry detection algorithm based on present deep learning methods, which could ensure accuracy and robustness for robotic strawberry harvesting in outdoor environments \citep{he2022detecting, yan2021real}. As mentioned before, the strawberry detection models for robotic harvesting need to focus on real-time speed, high precision as well as robustness in the field environment. Therefore, improving the model performance on in-field strawberry detection while ensuring the lightweight property of the detection model has become the focus of this study. 

The primary goal of this study was to develop an accurate and efficient strawberry detection model for robotic harvesting in the outdoor field environment. To achieve this goal, the specific following objectives of the study were pursued in this study:

\begin{enumerate}[ ]
            \item Adoption of the YOLOv5 structure with modifications: The study aimed at modifying the original YOLOv5 structure to improve the accuracy of strawberry detection in open-field conditions. Specifically, the modifications included replacing the C3 modules with C2f modules and introducing SPPFCSP blocks.
            \item Performance comparison with other YOLO models: The proposed model's performance (in terms of detection accuracy and inference speed)  was compared with several competitive models, namely YOLOv3-tiny, YOLOv5s, YOLOv5s-C2f, and YOLOv8s. 
        \end{enumerate}
By addressing these objectives, the study aimed at contributing to the advancement of robotic strawberry harvesting by developing an effective strawberry detection model that outperforms existing models in terms of accuracy and efficiency.

\section{Materials and method}
\subsection{Strawberry dataset}
\begin{figure}[!h]
  \centering
  \subfigure[]{
    \includegraphics[width = 7.5cm,height = 4.3cm]{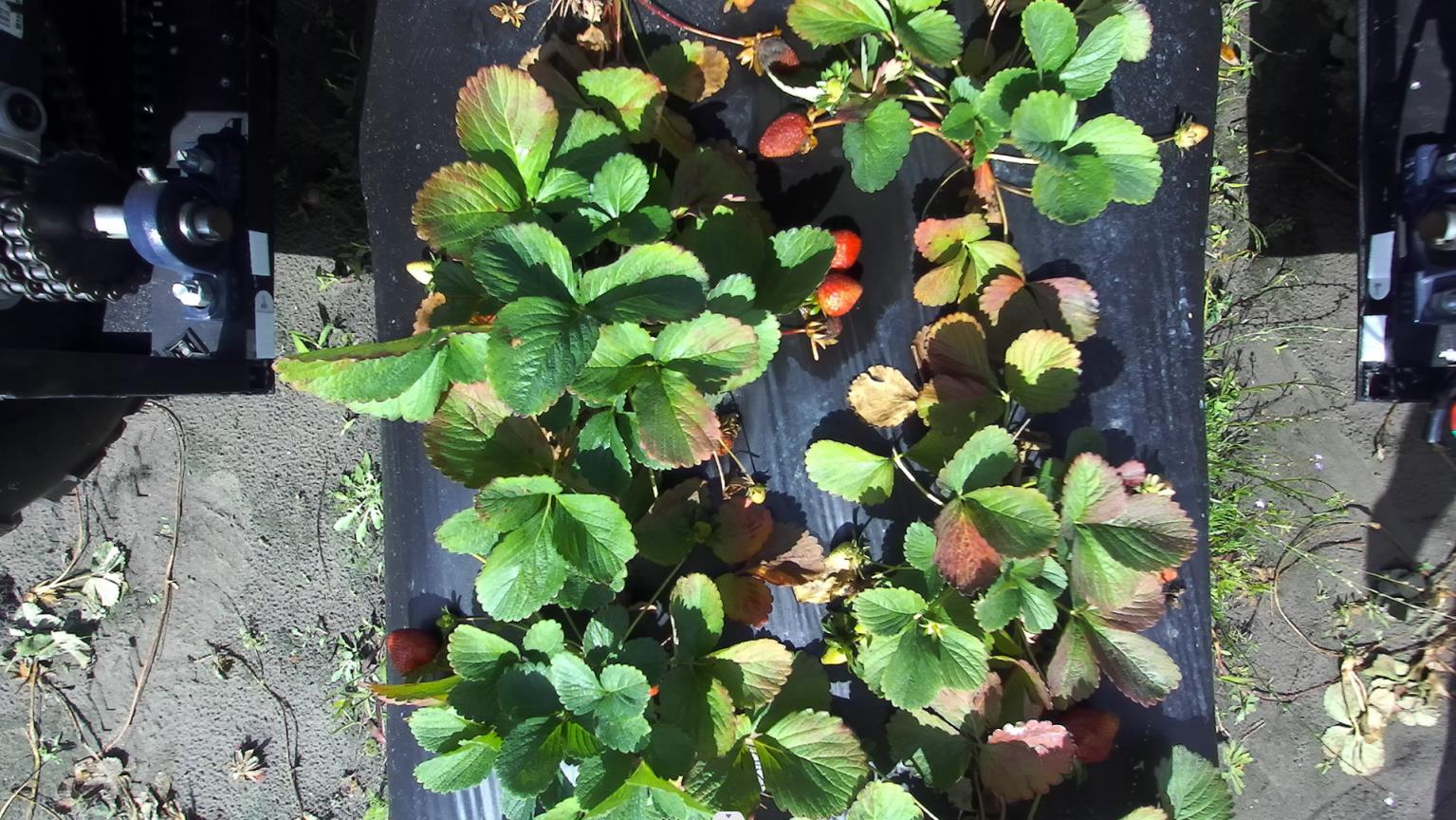} \label{f1a}
  }
  \subfigure[]{
  \includegraphics[width = 7.5cm,height = 4.3cm]{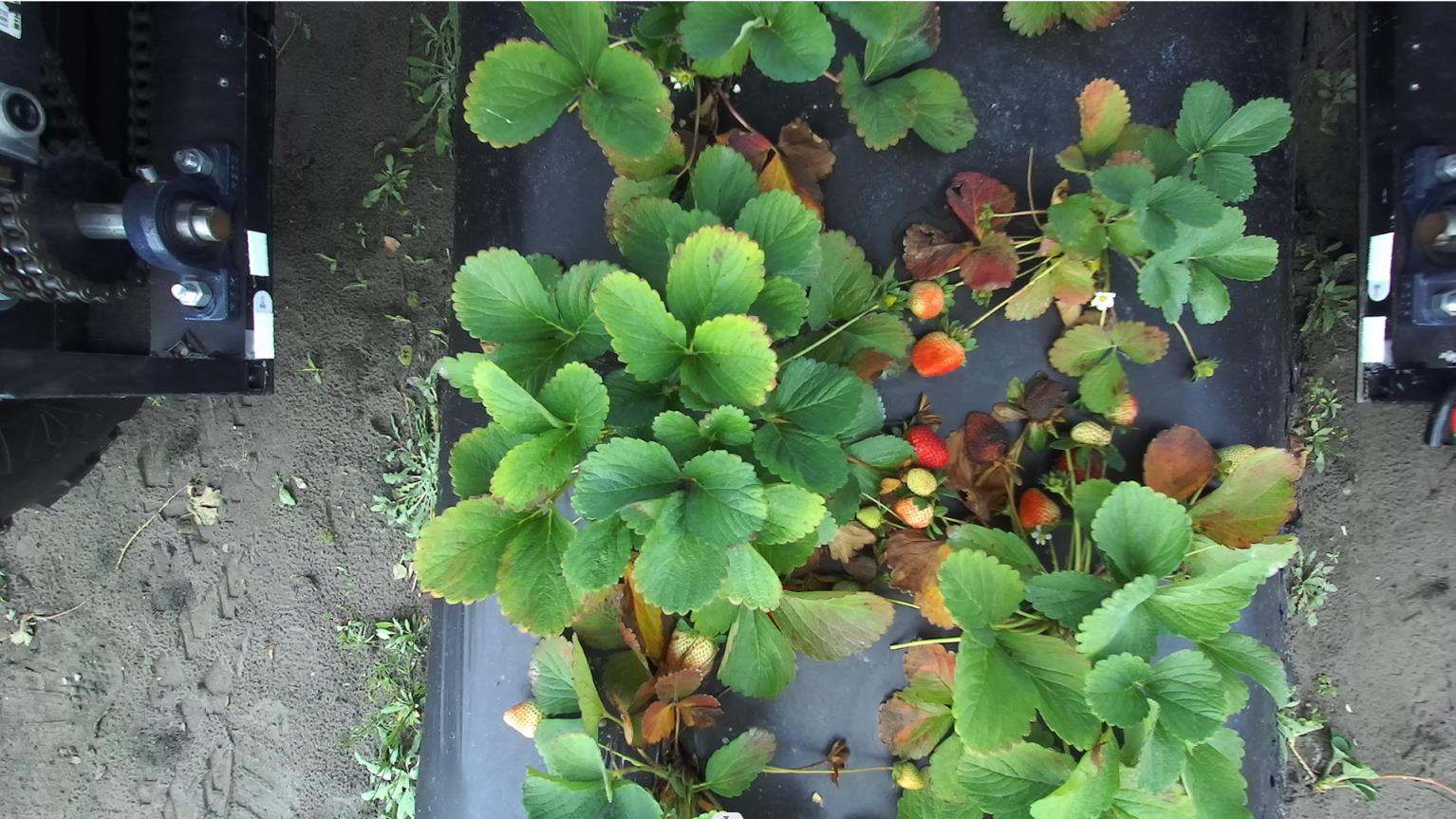} \label{f1b}
  }

  \caption{strawberry canopy image in different lighting conditions:a) sunny condition with shadow between the canopies; b) cloudy condition.}
  \label{f1}
\end{figure}

In this study, RGB images of strawberry canopies were collected in a commercial field near Orlando, Florida, USA. The width of strawberry ridges was ~1.3 m, and the space between each canopies was between 50 and 70 cm, allowing most strawberries to be completely or partially visible to the camera.  The images were collected twice a day, around 10:00 a.m. to noon and 2:00 – 5:00 p.m. from Feburary 5  to  February 10, 2020. A ZED2 stereo camera (Stereolabs Inc.,2019) was used for acquiring the canopy images at a resolution of 2208 × 1242 pixels. An imaging platform/cart was used to mount the camera and was manually pushed into the field to maintain an approximate camera height of 120 cm above the ground. Due to uneven ground in the field, the height and imaging angle of the camera varied slightly. Additionally, all the data collection was carried out in natural light including cloudy and sunny conditions to improve the generality of the deep learning model trained (\citep{gao2020multi}. In total, 1,540 RGB images were acquired in sunny (Figure \ref{f1a}) and cloudy conditions (Figure \ref{f1b}). There were more shadows observed in the canopies under the sunny condition, which created more diversity in training dataset, compared to the images under cloudy condition.

\subsection{Dataset annotation and augmentation}

\label{section 2.2.1}

\begin{figure}[!ht]
  \centering

     \subfigure[]{
    \includegraphics[width = 7cm,height = 3.8cm]{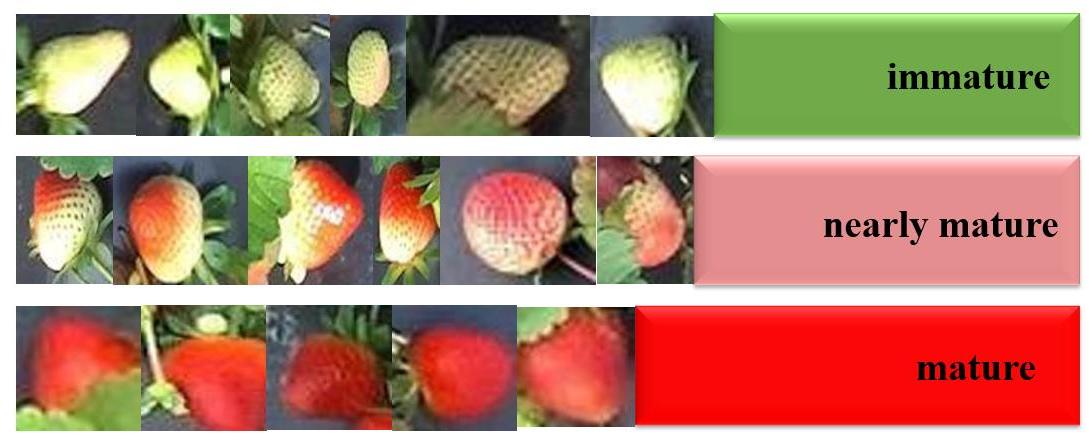} \label{f2a}
  }
  \subfigure[]{
  \includegraphics[width = 7cm,height = 4.6cm]{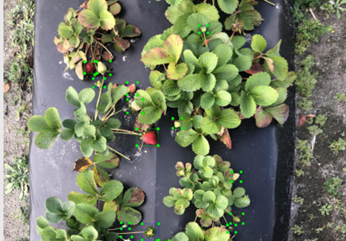} \label{f2b}
  }
 
  \caption{a)Example images showing strawberries in four maturity levels (immature, nearly mature, mature);b) example of labeled strawberries in the canopy.}
  \label{f2}
\end{figure}

Strawberries in the canopies undergo a series of physiological and biochemical changes during the growing season, leading to changes in fruit color and other characteristics unique to different maturity levels of strawberries \citep{barnes1976cell}. The implementation of three maturity categories including immature, nearly mature, and mature would significantly enhance the precision of identifying ripe strawberries, reducing the occurrence of false positives over the nearly mature groups during detection. Moreover, the inclusion of an additional category between immature and mature groups, specifically denoting strawberries in a nearly mature state, would furnish growers with crucial insights to strategically plan their subsequent harvesting endeavors. In this study, the maturity levels are classified into three categories; namely: immature, nearly mature, and mature (see Figure\ref{f2a}), which are described as follows:

 \begin{enumerate}[label= ]
            \item \textit{Immature stage} ‒  green or white strawberries (no red surface) 
            \item \textit{Nearly mature stage} – reddish strawberries (1/ 3 to 3/4 of the fruit surface area covered by red color) 
            \item \textit{Mature stage} – red ripe strawberries (red color over 3/4 of the surface area)
        \end{enumerate}

\begin{figure}[!h]
  \centering

     \subfigure[]{
    \includegraphics[width = 4.5cm,height = 4.5cm]{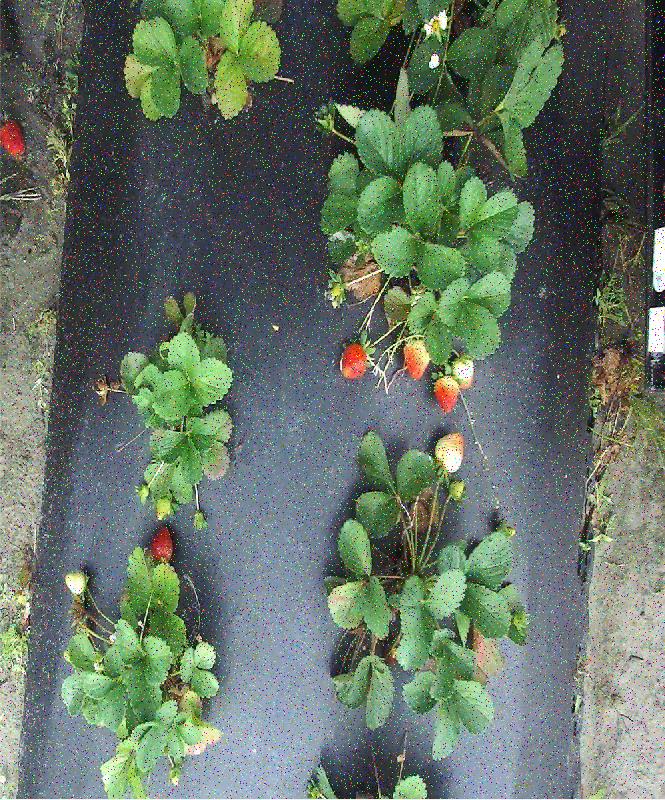} \label{f3a}
  }
  \subfigure[]{
  \includegraphics[width = 4.5cm,height = 4.5cm]{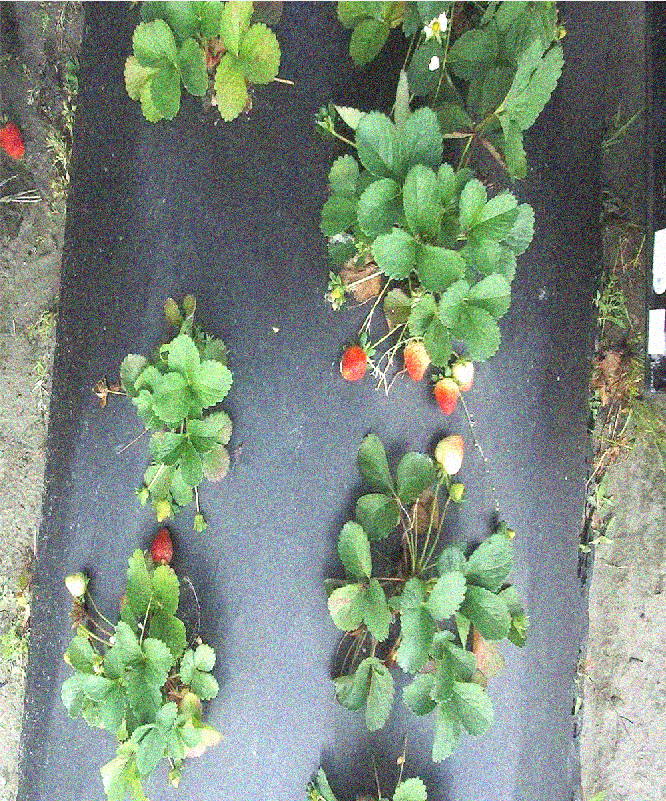} \label{f3b}
  }
  \subfigure[]{
  \includegraphics[width = 4.5cm,height = 4.5cm]{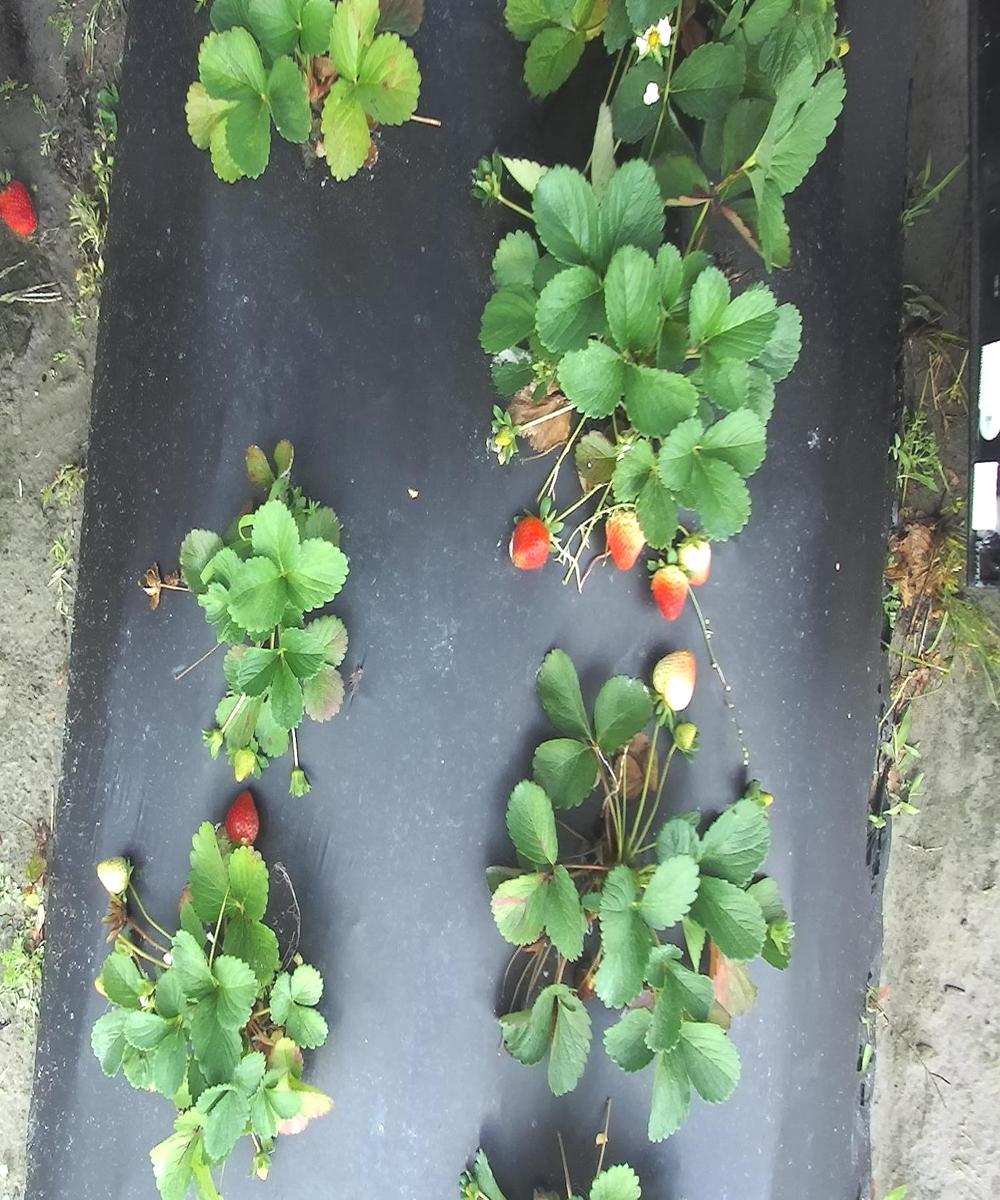} \label{f3c}
  }
  \subfigure[]{
  \includegraphics[width = 4.5cm,height = 4.5cm]{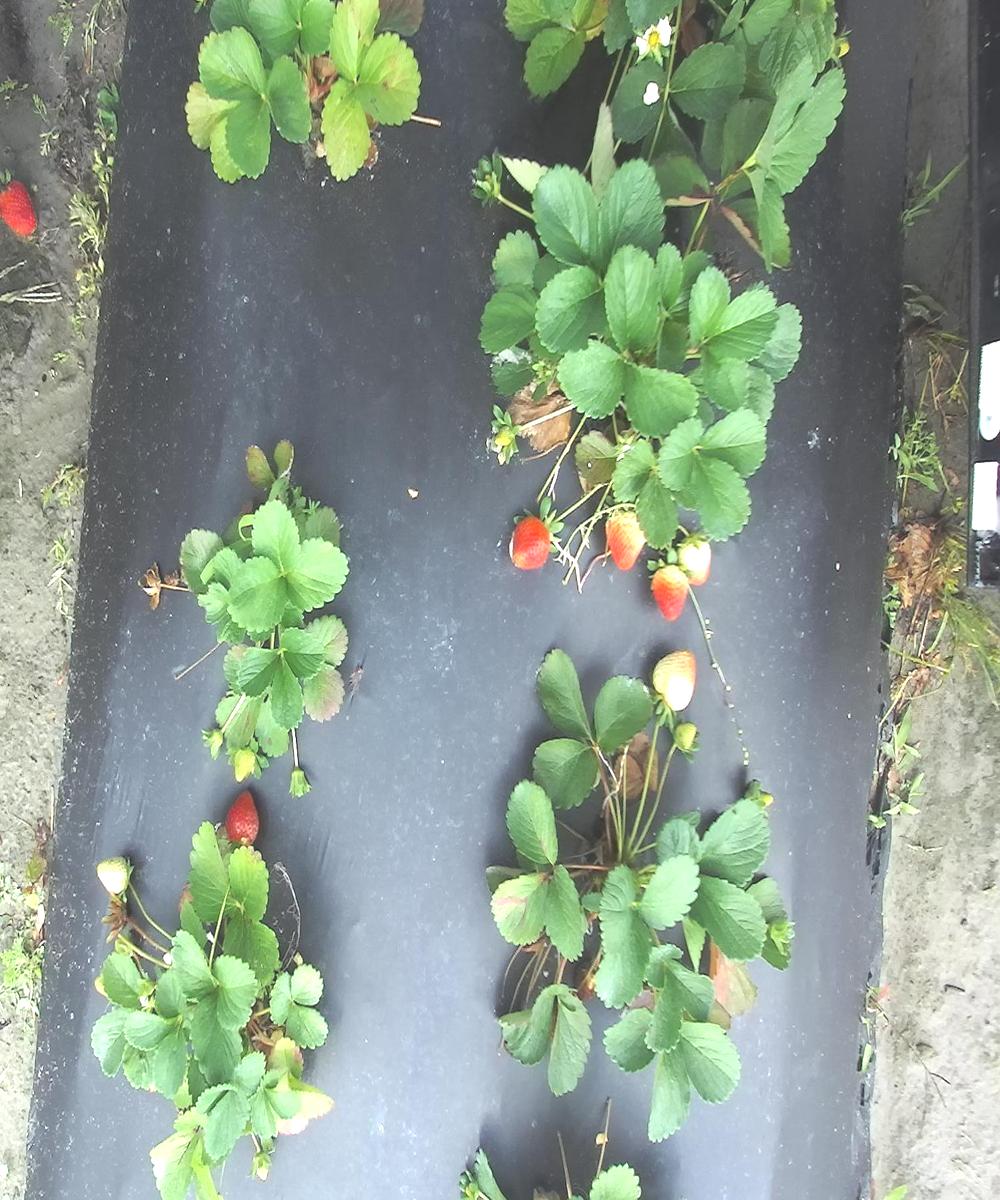} \label{f3d}
  }
  \subfigure[]{
  \includegraphics[width = 4.5cm,height = 4.5cm]{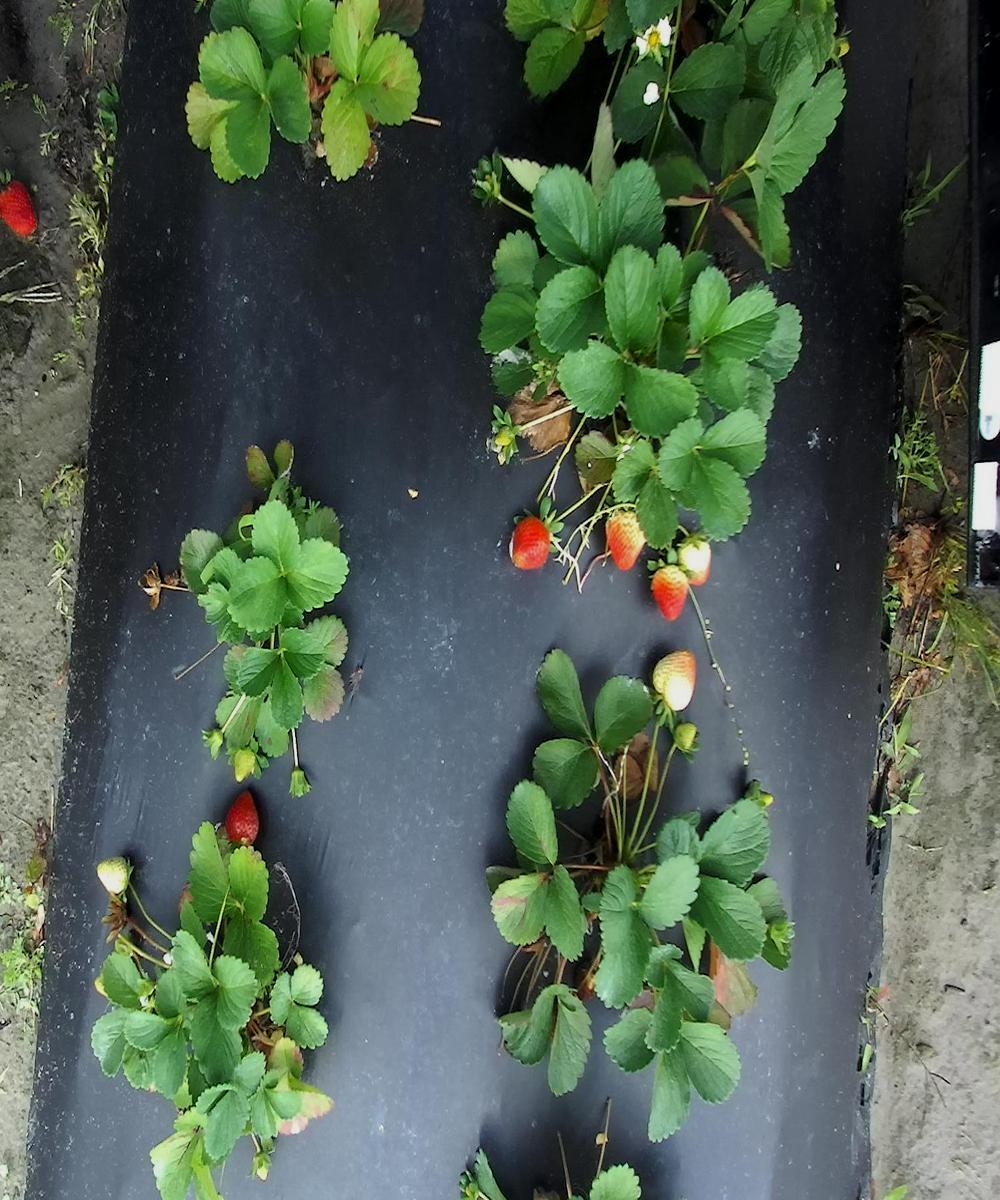} \label{f3e}
  }
  \subfigure[]{
  \includegraphics[width = 4.5cm,height = 4.5cm]{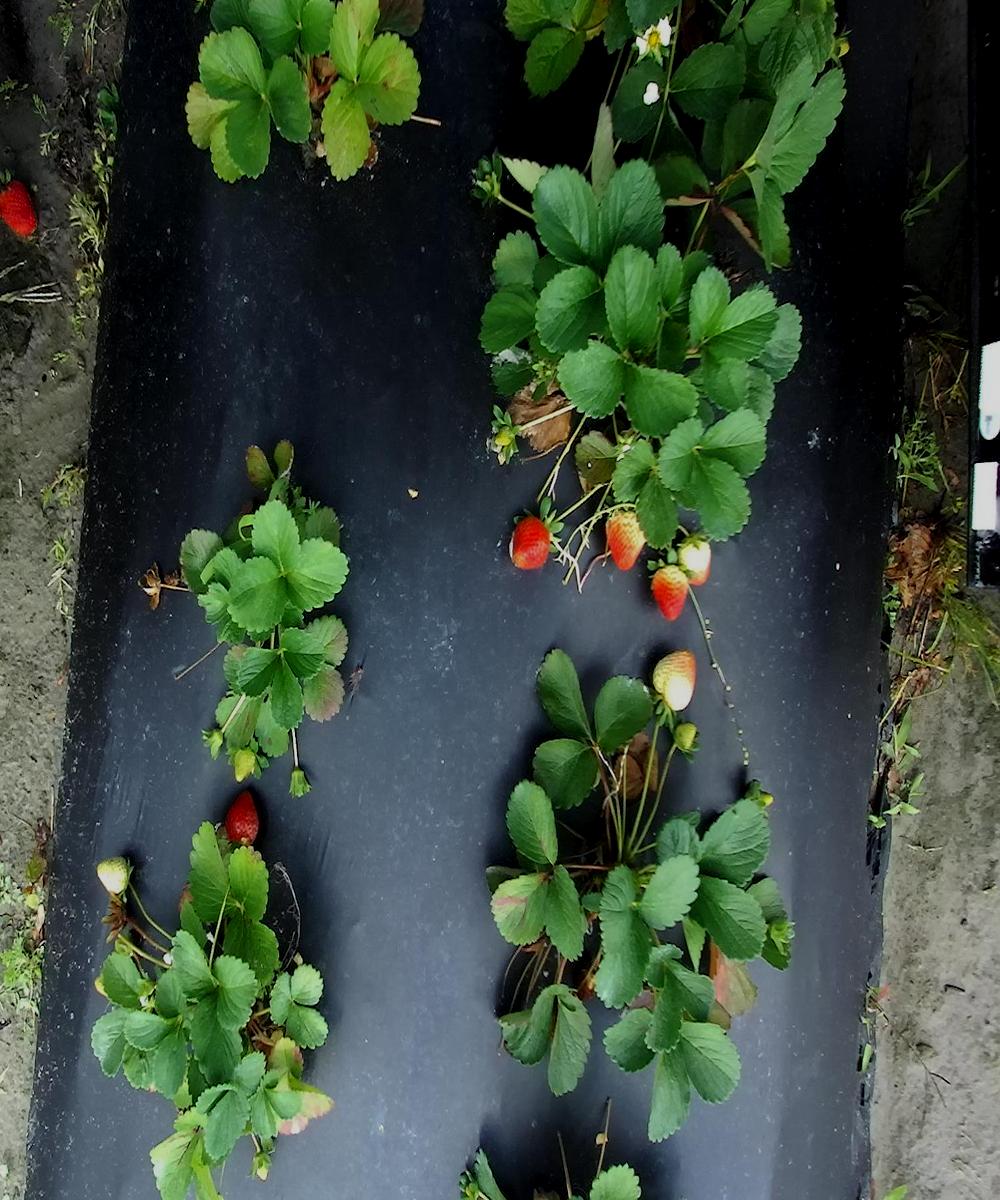} \label{f3f}
  }

  \caption{Strawberry canopy image after different augmentation methods: a)salted pepper noise; b)gaussian noise; c) brightness enhancement of 20; d) brightness enhancement of 40; e) brightness reduction of 20; brightness reduction of 40.}
  \label{f3}
\end{figure}

Since fruits in the strawberry canopies were in different growth stages, there were over 10,631 labeled objects in the immature group, 2,454 in the nearly mature group, and 6,617 in the mature group,  in the training dataset. The dataset is split into training and test in the ratio of 90:10. To train YOLO models for detecting target objects (strawberries with three different maturity levels), the location and maturity of individual strawberries in each canopy image were labeled using \textit{Labelimg}\citep{tzutalin2015labelimg} (Figure \ref{f2b}). In order to enhance the diversity of the image dataset including illumination conditions and noise levels, image augmentation methods were applied to the image dataset. The image augmentation techniques used included brightness enhancement, brightness reduction, and introduction of artificial noise using MATLAB (2022a, the MathWorks Inc., Natick, MA, USA). Salt pepper noise and Gaussian noise with a variance of 0.02 (Figure  \ref{f3a} and \ref{f3b}) were also added to the raw images. The brightness enhancement ( Figure \ref{f3c} and \ref{f3d}) was achieved by increasing raw pixel values by 20 and 40.  Similarly, The brightness reduction (Figure \ref{f3e} and \ref{f3f}) was achieved by decreasing pixel intensities by 20 and 40. The final training dataset (after these image augmentation techniques) used for model training included 7,935 images. The average number of strawberries in each image was approximately 15 with maturity ranging from immature stage to mature stage. Additionally,  mosaic data enhancement technique was applied in the input layer of YOLOv5, which combines four images into one by resizing each of four images, stitching all together, and random cutout of the stitched images \citep{Jocher_YOLOv5_by_Ultralytics_2020}. This data enhancement in the input layer of YOLOv5 can improve the data concentration and the imbalance of small, medium, and large target data during training (e.g., different sizes of strawberries in this study).

\subsection{Improvements of YOLOv5s Architecture}
YOLOv5 is a lightweight network with high computational speed while also providing high object detection accuracy and structure which could be modified for further adaptation to a specific problem. YOLOv5 includes five different models with varying network sizes: YOLOv5n, YOLOv5s, YOLOv5m, YOLOv5l, and YOLOv5l \citep{Jocher_YOLOv5_by_Ultralytics_2020}. The difference among these five versions of YOLOv5 includes depth multiple, width multiple, and ratio on the model size. These configurations would adjust the feature extraction modules and convolution kernel in fixed locations of the network that alters the total number of trainable parameters and Giga Floating-point Operations Per Second under a given computational environment \citep{Jocher_YOLOv5_by_Ultralytics_2020}.
Operating in open fields, robotic strawberry harvesters require a machine vision system with real-time detection capability and a lightweight network structure so that it can be executed on a light, mobile computational platform. Thus, the YOLOv5s, considering the accuracy and efficiency of these five models, was selected and modified to improve the overall performance on strawberry detection.

\subsubsection {Original YOLOv5s Architecture}
\begin{figure}[!h]
  \centering
  
    \includegraphics[width = 16cm,height = 5.2cm]{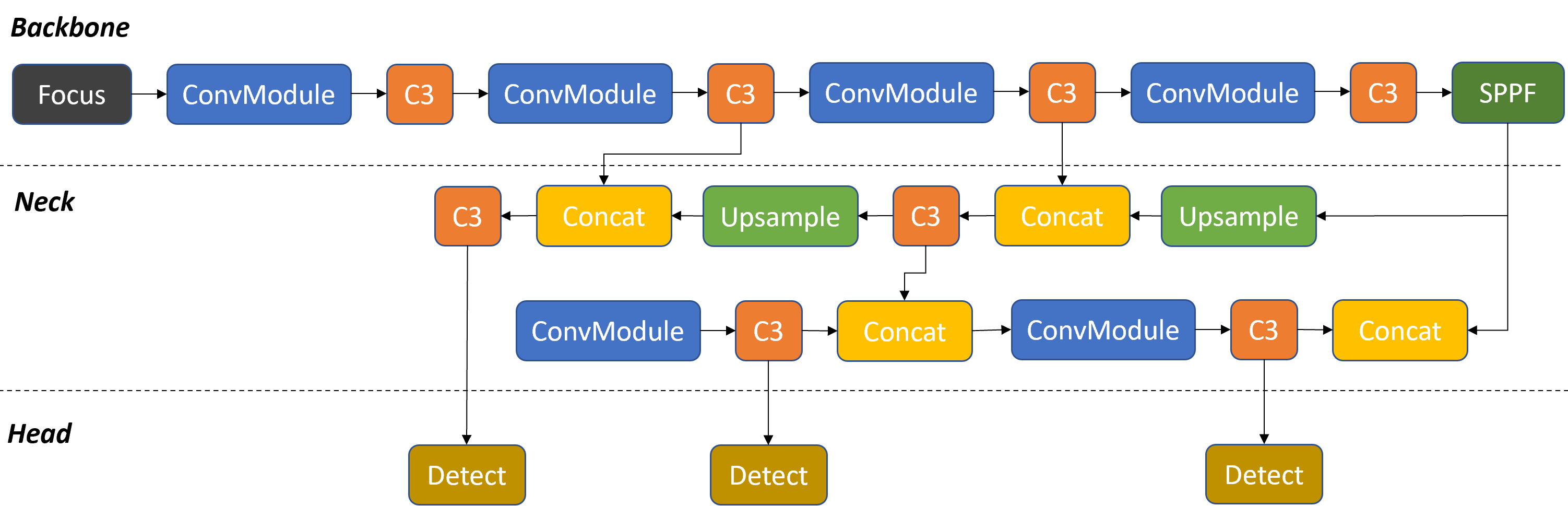} \label{fy51}

  \caption{YOLOv5 architecture including backbone, neck, and head from \cite{Jocher_YOLOv5_by_Ultralytics_2020}}
  \label{yolov5}
\end{figure}

YOLOv5 architecture (Figure \ref{yolov5}) consists of mainly three operating components: the backbone, neck, and head. The backbone network (CSPDarknet53) of YOLOv5 (Figure \ref{yolov5}, top) is used for extracting object  features from the input images with multi-modules of convolutions. The extracted feature map is in the sizes of 80 ×80, 40 ×40, and 20 ×20. The neck network (Feature Pyramid Networks and Path Aggregation Network) (Figure \ref{yolov5}, Middle) produces feature scale, and the head collects the features from the backbone. During this procedure, feature maps of different size generated by the backbone are fused to obtain more contextual information and reduce information loss. In the neck, a feature pyramid structure of the feature pyramid network (FPN) and path aggregation network (PANet) is adopted where the FPN structure is used to transfer strong semantic features from the top to the bottom feature map. At the same time, PANet is used to transfer strong location features from lower feature maps to higher feature maps. The head network (Figure \ref{yolov5}, Bottom) includes three detection layers with the relative output from the feature maps of 80×80, 40 ×40, and 20 ×20, which is applied to generate the output including bounding boxes and scores of the detected objects by applying the anchor boxes.

Although the original YOLOv5 models can realize strawberry detection, its architecture could be further improved for higher accuracy and robustness by modifying the backbone components from the latest YOLOv8.

\begin{figure}[!h]
  \centering
  \subfigure[]{
     \includegraphics[width = 10cm,height = 2cm]{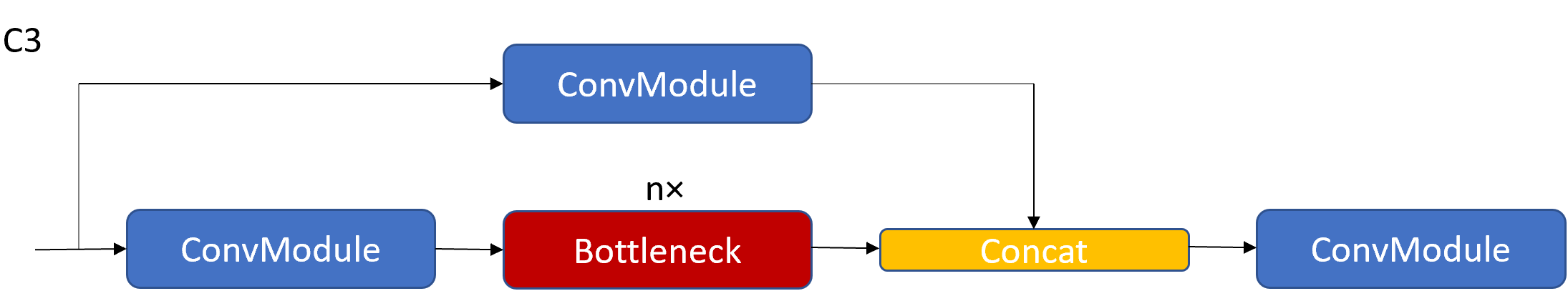} \label{C3}
  }

 \subfigure[]{
     \includegraphics[width = 10cm,height = 2.5cm]{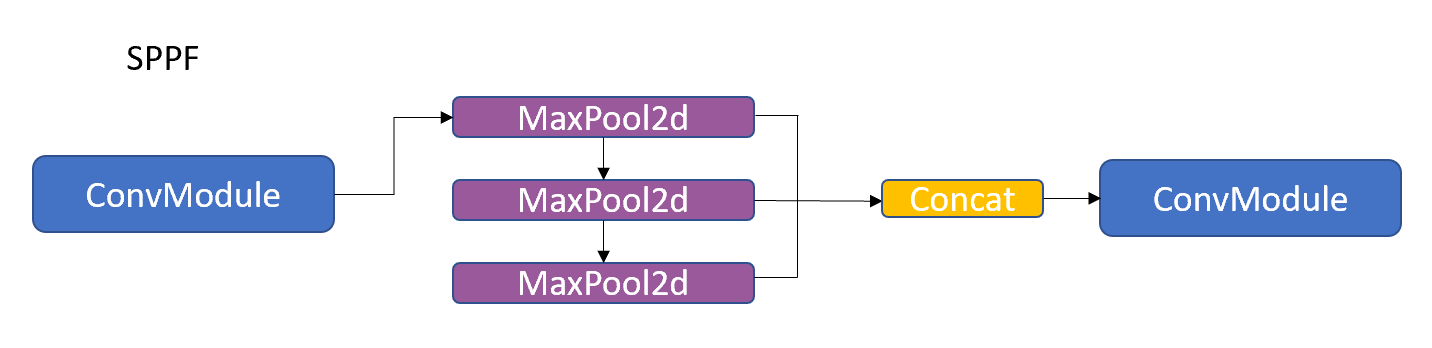} \label{s1}
  }

  \caption{Components of YOLOv5s: a) C3 module; b) SPPF module}

\end{figure}

The C3 modules are in the third, fifth, seventh, and ninth layers of backbone networks \citep{Jocher_YOLOv5_by_Ultralytics_2020}, which were Cross Stage Partial (CSP) Bottleneck with three convolution modules (Conv). The initial input inside the C3 module (Figure \ref{C3}) is divided into two branches, which leads to the number of channels in the feature map being halved by the Conv operation in each branch. The output feature map of the two branches is connected again through the Concat operation. The final feature map of C3 is generated by Conv. The advantage of C3 is used to improve the inference speed by reducing the size of the model while maintaining the expected performance of extracting useful features from the input image \citep{Jocher_YOLOv5_by_Ultralytics_2020}.

The final layer of the backbone network in YOLOv5 is Spatial Pyramid Pooling Fast (SPPF) (Figure \ref{s1}), which is applied for improving the receptive field by transferring feature maps of arbitrary size into feature vectors of fixed size, avoiding the problem of image distortion caused by image region clipping and scaling operation, and reducing the computational cost during detection \citep{he2015spatial, Jocher_YOLOv5_by_Ultralytics_2020}. The input feature map (size: 512 ×20×20) goes through the Conv layer first and then passes through three Maxpooling layers with sub-sampled information in the Concat operation. After final processing in the Conv module, the output feature map (size: 512 ×20×20) is obtained by the neck network through a convolution layer including 512 kernels.

\subsubsection{Modified Architecture in YOLOv5s}

\begin{figure}[!h]
  \centering

 \subfigure[]{
    \includegraphics[width = 13.5cm,height = 1.7cm]{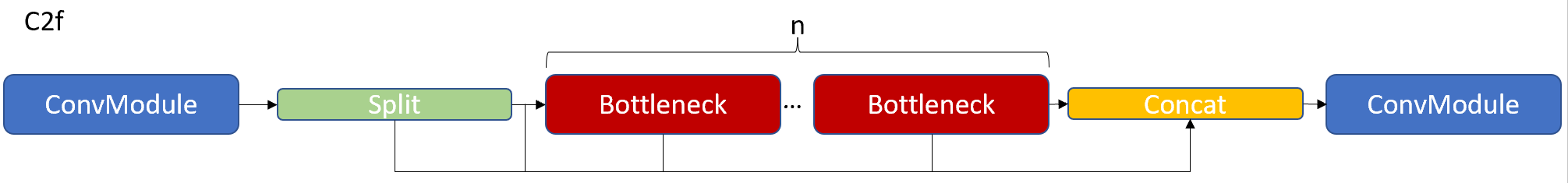} \label{C2f}
  }
  \subfigure[]{
      \includegraphics[width = 14cm,height = 3cm]{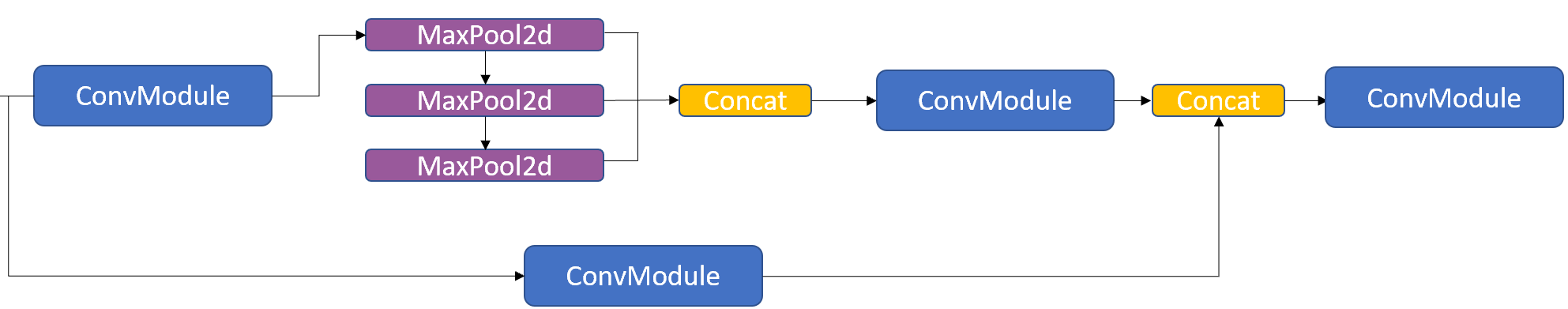}   \label{s2}

  }
    \caption{Componet of modified YOLOv5s: a) C2f module; b) SPPFCSP module}
  
\end{figure}

In the backbone of YOLOv5 architecture, C3 modules, could be improved structurally by adding the connection between CSP Bottleneck to the Concat operation, which enriched the analysis over feature map transferred from the target features. In this study, the C2f module is a new structure introduced from YOLOv8, which consists of two Conv modules with residual connection \citep{Jocher_YOLO_by_Ultralytics_2023} (Figure \ref{C2f}). C2f module is utilized for improving the performance of strawberry recognition in YOLOv5. In contrast to the C3 module in YOLOv5 (Figure \ref{C3}), the C2f module incorporates an additional operation on the split, involving multiple divisions from the Bottleneck connected to the Concat. This design leads to a richer gradient flow information while maintaining a lightweight overall network structure. Besides, the increase of the bridge branch from the Bottlenecks to the Concat operations in C2f benefits the flexibility of the feature map while keeping the C2f in a small parameter size. Therefore, four C3 modules in the backbone were replaced by four C2f modules in this study.

\begin{figure}[!h]
  \centering  
  
    \includegraphics[width = 16cm,height = 5cm]{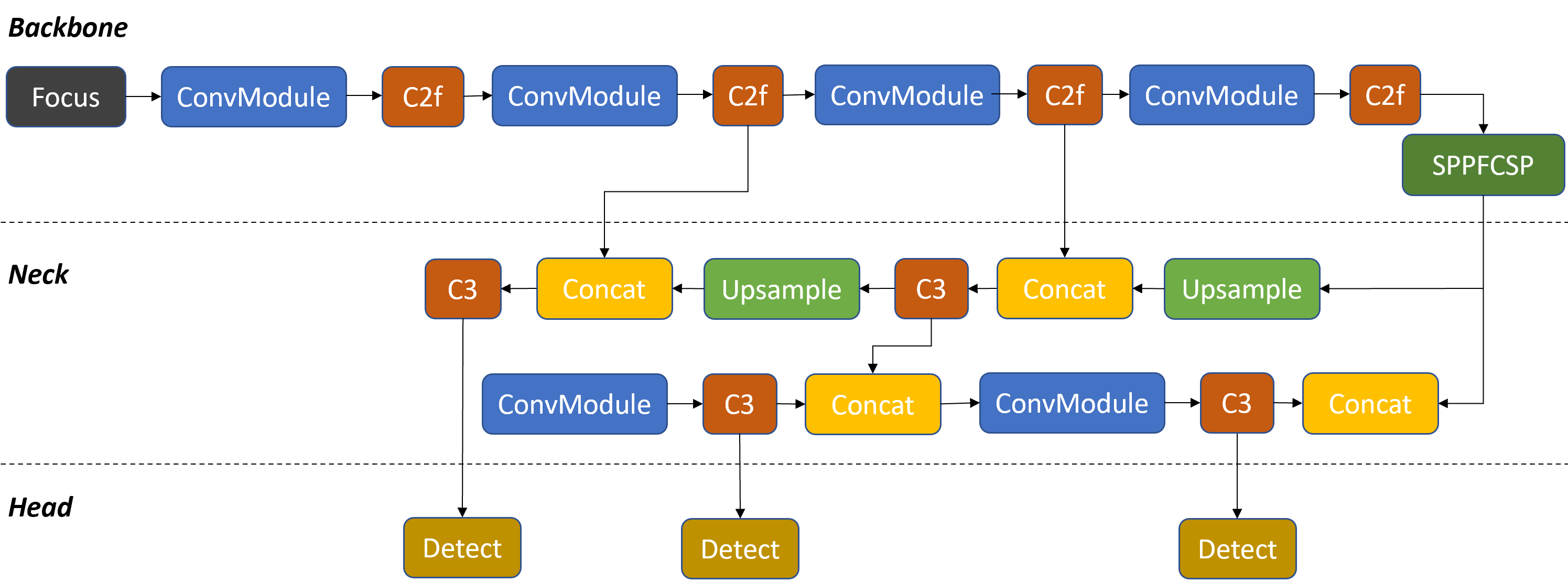}

  \caption{Improved YOLOV5 Architecture, modified by replacing C3 module and SPPF of the original YOLOv5 by C2f and SPPFCSP respectively }
  \label{c2f-spp}
  
\end{figure}

Over the last layer of YOLOv5, SPPF, the Cross Stage Partial Network (CSPNet) \citep{wang2019cspnet} was combined to optimize the computation speed, as shown in Figure \ref{s2}.  Firstly, the feature map is divided into two parts, one of which is for conventional processing, the other is for SPPF structure processing, and the two parts are combined through the Concat operation. The new structure of SPPF-CSPNet is similar to the SPPF used by YOLOv5 to increase the receptive field of a network. The input feature map with a size of 512 × 20 × 20 was obtained and subjected to three convolution operations. Max pooling operations with kernel size of 5, 9 and 13 were performed three times. Finally, the feature map with a size of 512 × 20 × 20 was obtained by combining the results with only 1 × 1 convolution operation without data pooling. With additional Concat operation from CSPNet, the final feature map will pass through the final Conv. By combining SPPF in YOLOv5 with CSPNet, the computational load in this block could be reduced by half while the recognition accuracy was expected to be improved \citep{wang2019cspnet}.  The structure of the improved design of the strawberry detection network is shown in Figure \ref{c2f-spp} where the part of the neck and head network was kept in the same structure as YOLOv5. The modified YOLOv5s in this study was neamed as YOLOv5s-Straw.

\subsection{Network training}
The  proposed model was trained on  the  Alienware  X15  laptop (Intel(R) with Core(TM) I7- 1180@2.3GHz  CPU, 32GB@3200Mhz memory, NVIDIA GeFore RTX 3070 GPU, and 24 GB GUP memory) on Windows 11 operating system. Python was employed to program the training and test procedure for strawberry detection in Pytorch framework utilizing CUDA, cuDNN, and OpenCV.
In this study, the modified YOLOv5s was trained in 100 epochs using stochastic gradient descent (SDG) learning method. The training process was set to terminate automatically while there was no observed improvement in the mean average precision based on the validation dataset in 20 training epochs (Patient set to 20). The batch size during the training process was set to 16 and a learning rate of 0.01 was used. The momentum was set to 0.937 and the weight decay rate was 0.0005. The initial vector and IoU (intersection over union) threshold were all set to 0.01. The enhancement coefficients for input images in hue (H), saturation (S), and lightness (V) were set to 0.015, 0.7, and 0.4 during the training process, respectively. The best and the last weight files were kept once the training process was completed. The final detection results included the location of bounding boxes with strawberry in three maturity levels and the probability of belonging to a particular maturity category.

Additionally, four light-weight detection networks, YOLOv3-tiny \citep{redmon2018yolov3}, YOLOv5s \citep{Jocher_YOLOv5_by_Ultralytics_2020}, YOLOv5s with C2f module (YOLOv5s-C2f), and YOLOv8s \citep{Jocher_YOLO_by_Ultralytics_2023}, were trained to compare the performance of the improved YOLOv5s model in detecting strawberries with the same achieved by these models.

\subsection{Performance assessment}
The performance of all the five models mentioned above in detecting strawberries was assessed using Precision(P), Recall (R), Average Precision(AP), mean Average Precision(mAP) with an intersection-of-union(IoU) over 50\%, and inference speed, in which mAP was a key metric to evaluate the overall performance of a model. The equations used to calculate these measures are as follows:
\begin{equation}
P = \frac{TP}{TP+FP}  
\end{equation}
\begin{equation}
R =\frac{TP}{TP+FN} 
\end{equation}

\begin{equation}
  mAP = \frac{1}{n} \sum_{k=1}^{k=n} {AP_k}
\end{equation}
\begin{equation}
   IoU = \frac{Area(I)}{Area(U)} 
\end{equation}

where \textit{TP} is the number of true positive strawberries detected, \textit{FP} is the number of false strawberries detected, and FN is the number of strawberries falsely not detected as strawberries; \textit{$AP_{k}$} represents the \textit{AP} of class \textit{k}; \textit{n} represents the number of classes. The \textit{IoU} threshold was set at 0.5, meaning that the result is true positive if the IoU between the predicted result and ground truth bounding boxes was greater than 0.5. 

\section{Results}
\subsection{Strawberry detection with proposed networks}
A test dataset of 140 images was used to assess the performance of the improved YOLOv5s model in detecting strawberries in three different maturity levels. The test images included a total of 2,201 strawberries, among which there were 1,234 immature fruit, 292 nearly mature fruit, and 665 mature fruit. The summary of the evaluation matrices are shown in Table \ref{t2}, which showed that the average precision (AP), precision, and recall on the mature class are 86.6\%, 73.2\%, and 89.3\%, respectively. The same for the nearly mature class of strawberries were 73.5\%, 62.0\%, and 76.5\% respectively and for the immature class were 82. 1\%, 74.4\%, and 80.3\%, respectively.  The mean average performance of the proposed network was 80.7\%.

\begin{table}[!h]
\caption {strawberry detection on test dataset}
\center
\begin{tabular}{c c c c c}\hline
Maturity&  Number & Precision & Recall & AP$^{a}$(\%) \\ \hline
immature &1234&74.4&80.3&82.1\\ 
nearly mature&292&62.0&76.4&73.5\\ 
mature&665&73.2&89.3&86.6\\ 
 \hline
\multicolumn{5}{l}{$^{a}$AP refers to  Average Precision.}\\
\end{tabular}\\
\begin{flushleft}

\label{t2}
\end{flushleft}
\end{table}

\begin{figure}[!h]
  \centering
  \subfigure[]{
    \includegraphics[width = 4cm,height = 4cm]{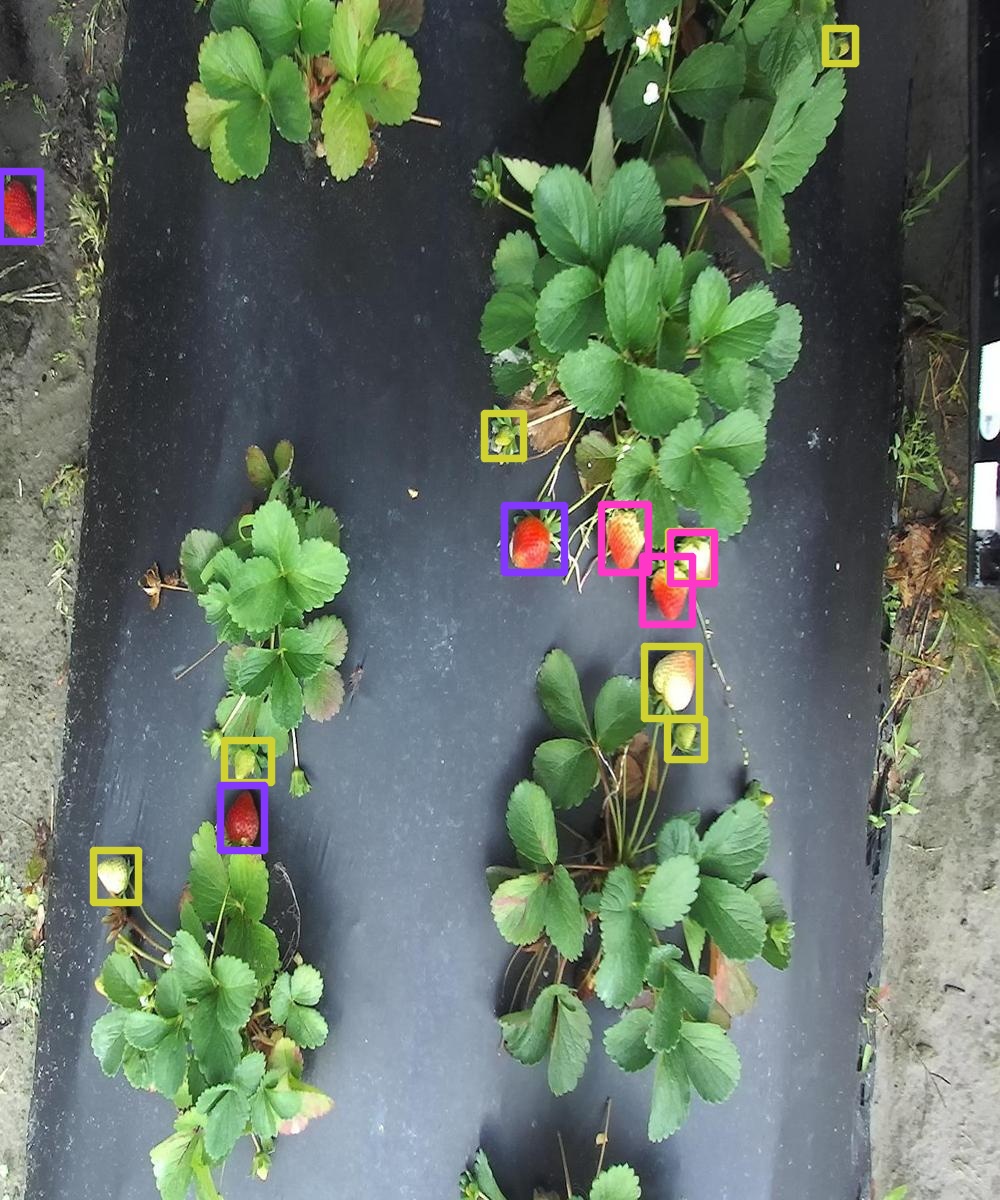} \label{testa}
  }
  \subfigure[]{
  \includegraphics[width = 4cm,height = 4cm]{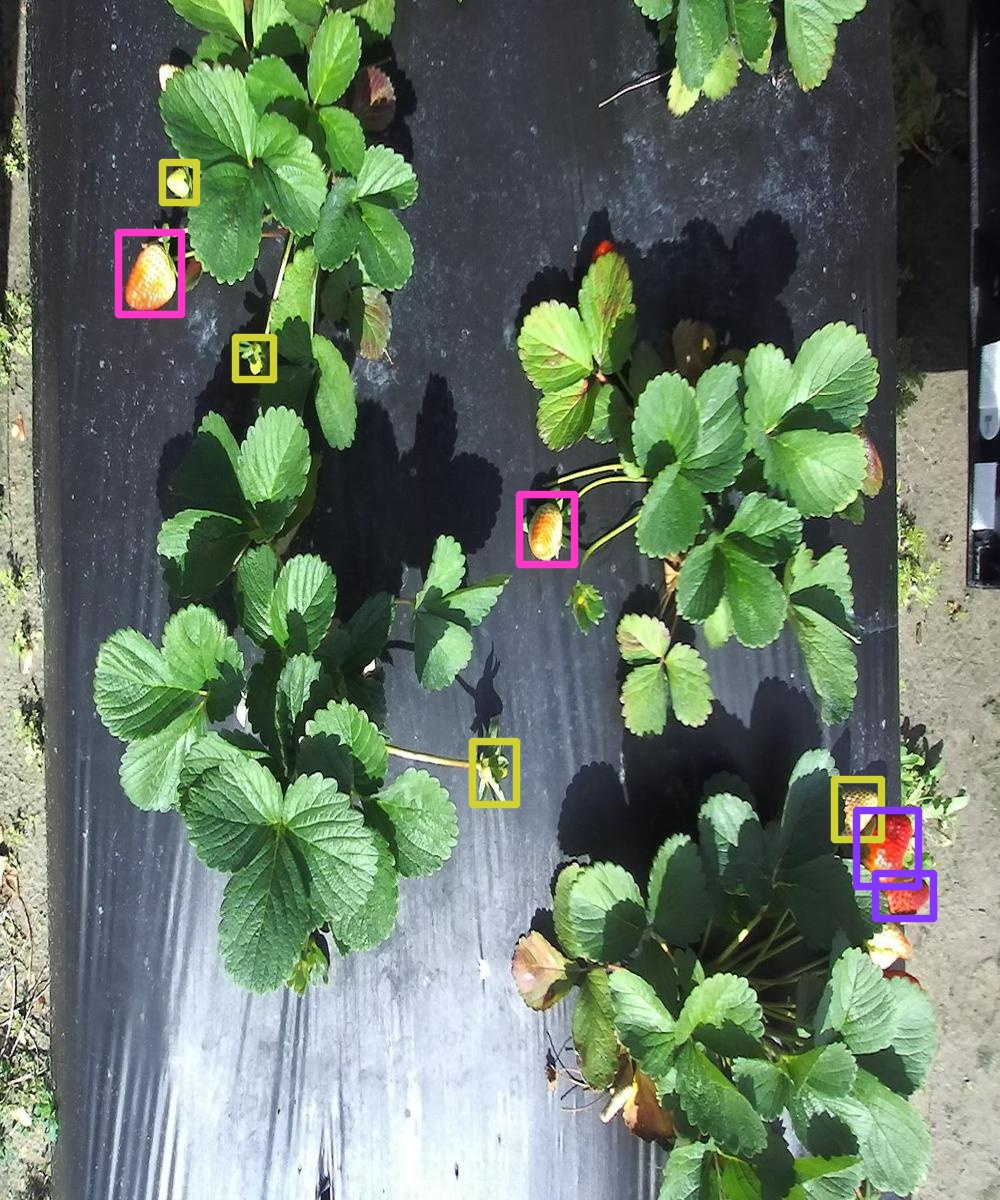} \label{testb}
  }
   \subfigure[]{
  \includegraphics[width = 4cm,height = 4cm]{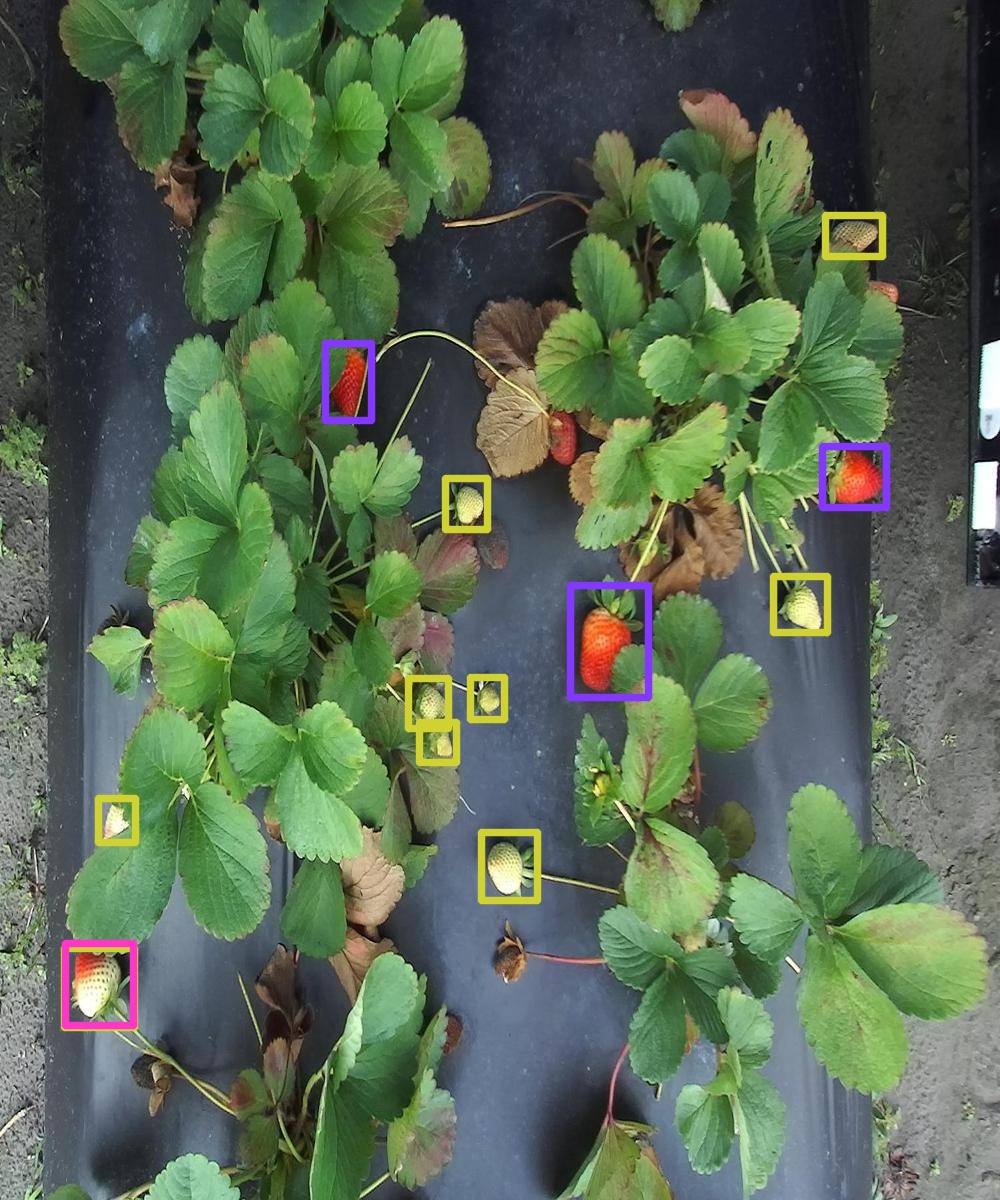} \label{testc}
  }
   \subfigure[]{
  \includegraphics[width = 4cm,height = 4cm]{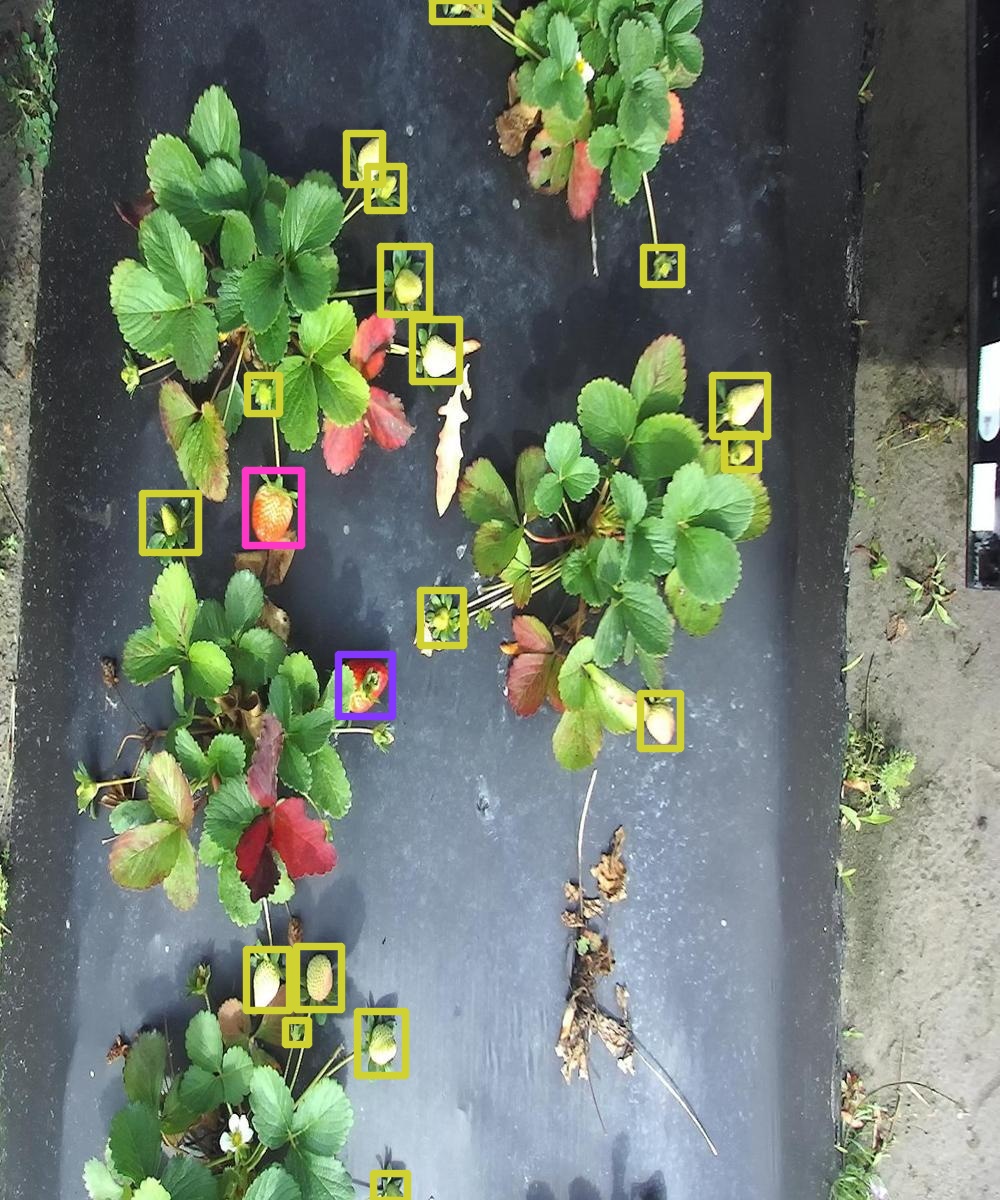} \label{testd}
  }
   \subfigure[]{
  \includegraphics[width = 4cm,height = 4cm]{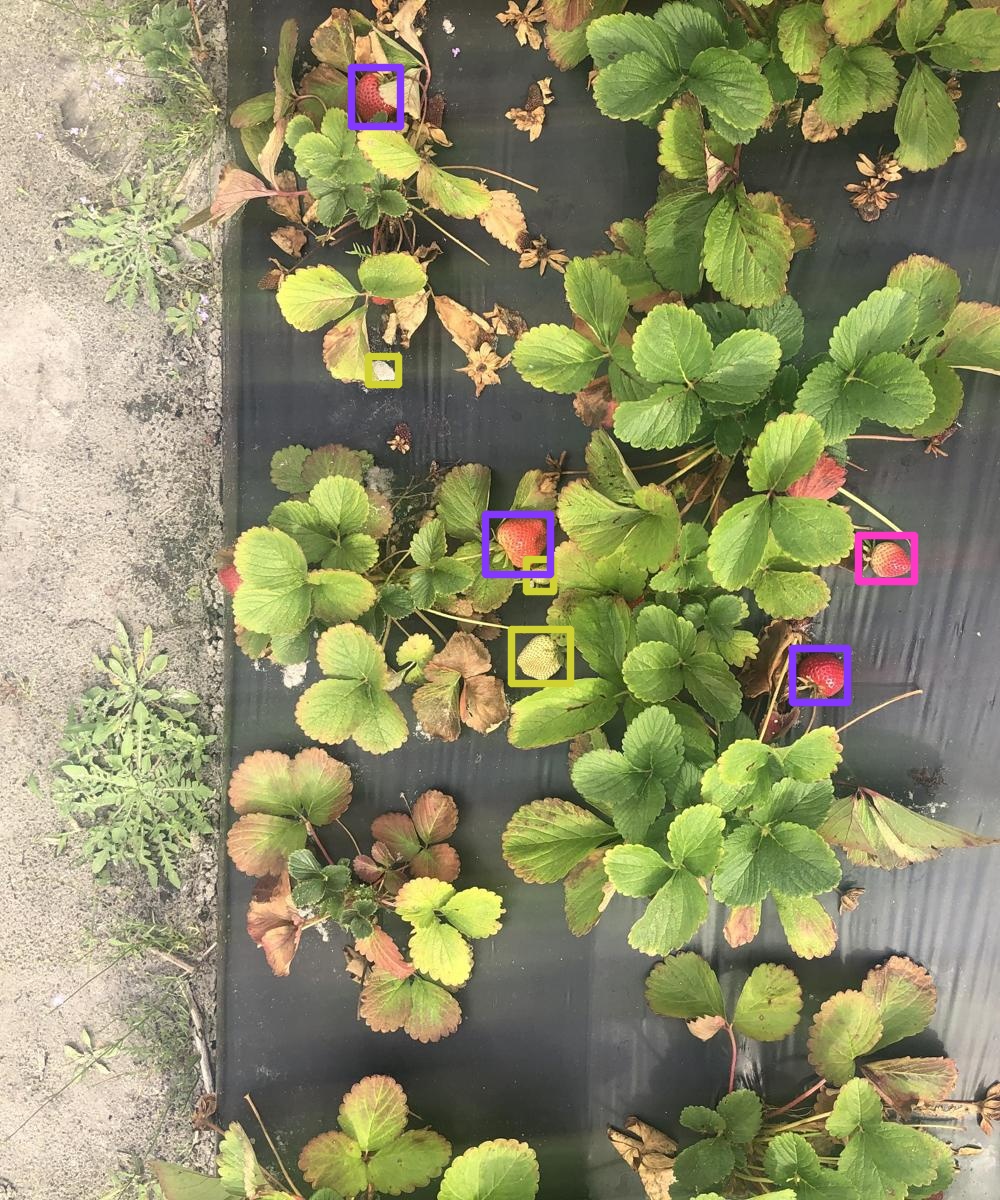} \label{teste}
  }
   \subfigure[]{
  \includegraphics[width = 4cm,height = 4cm]{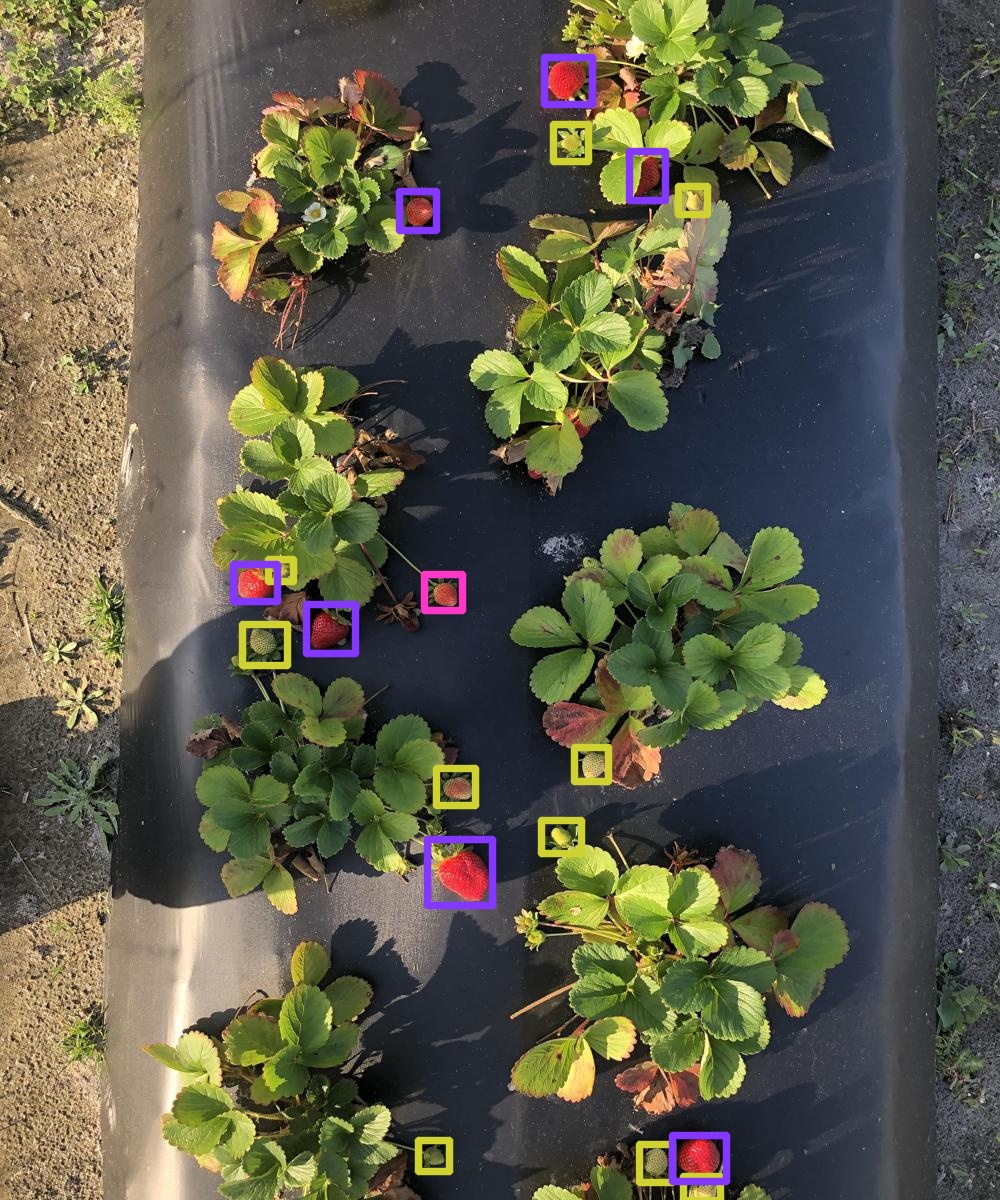} \label{testf}
  }

  \caption{strawberry canopy detection in different lighting conditions: a) c) cloudy condition in the morning; b) d) sunning condition in the morning; e) cloudy condition in the afternoon ;f) cloudy condition in the afternoon.}
  \label{e1}
\end{figure}
The example results presented in Figure \ref{e1} showcase strawberries at multiple maturity levels captured in different environmental conditions. In the images, the red bounding boxes represent the 'immature' group, the orange bounding boxes represent the 'nearly-mature' group, and the pink bounding boxes represent the 'mature' group. The proposed strawberry detection model demonstrates adaptability to various weather conditions, including sunny scenarios illustrated in Figure \ref{testa}, \ref{testc}, and \ref{teste}, as well as cloudy conditions depicted in Figure \ref{testb}, \ref{testd}, and \ref{testf}. Despite the presence of shadows and reverberation lights in the images captured under sunny conditions, the target fruits are still detected with a reasonable level of accuracy. Furthermore, the three varieties of strawberry images captured during various time of the day were effectively recognized, each potentially subject to distinct light conditions, as illustrated in Figure 8(c) captured in the morning and Figure 8(f) taken in the afternoon.

\subsection{Comparison of different strawberry detection models}
\label{section 3.2}

To comprehensively evaluate the performance of the strawberry detection method proposed in this study, a comparative study was conducted with four detection models with light-weight architectures: YOLOv3-tiny, YOLOv5s, YOLOv5-C2f, and YOLOv8s (the latest YOLO model). The specifications of each model used in the comparison are detailed in Table \ref{t1}. Performance measures including average precision for each maturity group, mean average precision, and inference speed per image were analyzed and presented in Table \ref{t6}.

The proposed YOLOv5s-Straw model in this study consisted of 159 layers and only 8.8$\times$10$^{6}$ parameters. It exhibits a Giga Floating-point Operations rate of 20.4 per second, with a compact size of 18.3 Mb. These characteristics highlight its lightweight properties, which have been achieved through the modification of the C2f modules and the introduction of SPPFCSP blocks. Comparatively, the YOLOv3-tiny model has a size of 33.4 Mb, YOLOv5s has a size of 14.0 Mb, YOLOv5s-C2f has a size of 17.0 Mb, and YOLOv8s has a size of 21.5 Mb. The YOLOv5s-Straw model demonstrates better performance than the original YOLOv5s structure and the YOLOv8s model, while maintaining a moderate parameter count and size between YOLOv5s and YOLOv8s. These results highlight the effectiveness and efficiency of the proposed YOLOv5s-Straw model for strawberry detection, showcasing its superiority over other models while maintaining a reasonable model size and parameter count.

\begin{table}[!ht]
\caption {Comparison on model specifications}
\center
\begin{tabular}{c c c c c}\hline
model&  number of Layers & Parameter & GFLOPS$^{a}$ & size(\textbf{\textit{MB}}) \\ \hline
YOLOv3-tiny&49&8.6$\times$10$^{6}$&13.0&33.4\\ 
YOLOv5s&157&7.0$\times$10$^{6}$&15.8&14.0\\ 
YOLOv5s-C2f&149&8.2$\times$10$^{6}$&19.5&17.0\\ 
YOLOv8s&168&11.1$\times$10$^{6}$&26.4&21.5\\
\makecell{YOLOv5s-Straw\\(\textbf{ours})}&159&9.4$\times$10$^{6}$&20.4&18.3 \\
 \hline
 \multicolumn{5}{l}{$^{a}$GFLOPS refers to Giga Floating-point Operations Per Second.}\\

\end{tabular}\\
\begin{flushleft}

\label{t1}
\end{flushleft}
\end{table}
\begin{table}[!ht]
\caption{Model comparison using the test set.}
\center
\begin{tabular}{c c c c c c c}\hline
\makecell{\multirow{2}{*}{Network}}& \multicolumn{3}{c}{AP$^{a}$($\%$)} & \multirow{2}{*}{\makecell{\textit{mAP}$^{b}$($\%$)\\ \textit{(IoU = 0.5)}}} & \multirow{2}{*}{\makecell{Inference speed \\per image (ms)$^{c}$}}\\ 

 &immature & mature & nearly mature & \\
\hline
YOLOv3-tiny&71.5 & 81.6& 67.2&73.4&17.8\\
YOLOv5s &79.9&84.1&69.5&77.8&\textbf{15.8}\\
YOLOv5-C2f&81.6&85.6&72.2&79.8&17.7\\
YOLOv8s&79.8&82.9&\textbf{75.1}& 79.3&21.0\\
\makecell{YOLOv5s-Straw\\ (\textbf{ours})}& \textbf{82.1}& \textbf{86.6}&73.5& \textbf{80.7}&18.1\\ 
\hline
\multicolumn{5}{l}{$^{a}$AP refers to  Average Precision.}\\
\multicolumn{5}{l}{$^{b}$mAP refers to mean Average Precision.}\\
\multicolumn{5}{l}{$^{c}$inference speed may vary under different computational environment.}\\
\end{tabular}

\label{t6}

\end{table}

Table \ref{t6} presents clear evidence of the superior performance of YOLOv5s-Straw, with a mean average precision (mAP) of 80.7\%. This results showed that the proposed model outperformed YOLOv3-tiny, YOLOv5s, YOLOv5s-C2f, and YOLOv8s by a margin of 7.3\%, 2.9\%, 0.9\%, and 1.4\%, respectively. Though marginal in some cases, these results establish the proposed method as the most effective for strawberry detection compared to the other models generally used for faster and more accurate object detection. Furthermore, our method demonstrated the highest average precision (AP) in detecting both immature and mature strawberries, achieving AP values of 82. 1\% and 86.6\%, respectively. These values surpass the performance of YOLOv8s by 2.3\% and 5.7\%, respectively.

The proposed model also exhibited substantial enhancements when compared to the original YOLOv5s, which has an mAP of 77.8\%. Notably, our model shows significant improvements in detecting strawberries at the immature stage (+3.2\%), nearly mature stage (+2.5\%), and mature stage (+4.0\%). However, the recognition rate for the nearly mature group is slightly lower by 1.6\% compared to YOLOv8s. Regarding computational speed, our model achieved an inference time of 18. 1 ms per image, while YOLOv5s demonstrates the fastest inference speed at 15.8 ms per image, which is 3.5 ms faster than YOLOv8s.

In summary, the YOLOv5s-Straw model proposed in this study not only achieved the highest mAP among the real-time detection models tested, but it also demonstrated superior performance in terms of AP for both immature and mature groups. Although our model has a slightly slower recognition speed compared to YOLOv5s, the speed remains practically applicable for real-time strawberry detection in robotic harvesting systems.

\section{Discussion}
Contrary to numerous other fruit crops like apples, tomatoes, and citrus fruits, strawberries cultivated in open-field settings demonstrate a distinct attribute: their canopies encompass fruits at diverse stages of maturity, encompassing flowering to fully ripe fruit. This variation in maturity levels necessitates the development of a vision system specifically optimized for detecting fruits that are ready for harvesting, while also providing accurate locations of immature fruits that should be left intact and avoided during the picking process. Furthermore, the outdoor strawberry fields often include narrower plant beds, and the fruits are softer, imposing a greater demand for precise detection to prevent any potential brush or damage during robotic picking.  The detection of strawberries for robotic harvesting in controlled environments such as greenhouses and polytunnels has been extensively studied \citep{defterli2016review,ge2019fruit}. In such indoor environments, with relatively stable lighting conditions, strawberry fruits are often arranged on two sides of plantation tables or flowerpots, allowing for easier separation from the leaves. However, in the context of open-field farming, detecting strawberries becomes more challenging due to variable lighting conditions and the presence of more complex canopy structures. Fruits are randomly distributed throughout the canopy, making their detection more intricate and demanding.

Given the ridge-planted structure of the strawberry canopy in the open-field environment (as depicted in Figure \ref{f1}), this study employed a perpendicular camera view to the plant bed during image acquisition in order to cover the wide field of view across the canopy. In this study, strawberries were classified into three distinct groups: mature, nearly mature, and immature. This classification approach has the potential to not only enhance the accuracy of mature fruit detection but also facilitate mapping and estimation of the volume of nearly mature fruit. Such information can be valuable for growers in planning subsequent passes of robotic picking. 

In this study, YOLO detection models with lightweight network structures were selected to compare the performance of the proposed model against their performances. In Section \ref{section 3.2}, it was notable that the detection results achieved by two original versions of YOLO models (YOLOv5s and YOLOv8s) were good with mean Average Precision (mAP) of 77.8\% and 79.3\%, respectively. The observed results showed that YOLOv8s had a slight advantage in recognizing strawberries in nearly mature class (AP: 75. 1\%) over YOLOv5s while performing similarly in the immature group with YOLOv5s. As mentioned previously in section 2.2, the number of labels in the immature and mature classes were nearly 4.5 and 3.7 times more than those in nearly mature class, which indicated the YOLOv5s might have better training performance with smaller dataset than that of the latest model,YOLOv8s. Specifically,  the Average Precision (AP) of YOLOv5s in mature class, as the most important target for robotic harvesting, was 84.1\%, which was 1.3\% higher than that of YOLOv8s. Moreover, the inference speed of YOLOv5s was only 87.2\% of that of YOLOv8 due to the lower number of trainable parameters (7.0$\times$10$^{6}$ vs11.1$\times$10$^{6}$) and lighter size (14.0 Mb vs 21.5 Mb) (Table \ref{t1}). These results indicated that YOLOv5s is more suitable for strawberry recognition for robotic harvesting than YOLOv8s. As YOLOv5s is computationally lighter and performs similarly to YOLOv8s in detecting strawberries in different maturity levels, the proposed study focused on further improving the model architecture. First, the strawberry detection performance in mAP of YOLOv5s-C2f, based on YOLOv5s with only adding C2f modules, achieved an overall improvement of 1.7\% over YOLOv5s in immature class, 1.5\% in mature class, and 2.7\% in nearly mature class while the processing time per image was improved by 12.0\%, from 17.7 ms to 15.8 ms. Ultimately, although there was slight increase in the size and number of trainable parameters in the final adaptation of YOLOv5s in our study over YOLOv5s-C2f because of the introduction of the new blocks of SSP with CSPNet, the proposed model achieved the best mAP of 80.7\% based on the test dataset while achieving a good inference speed of 18.1ms per image.

 There are few studies on open-field strawberry detection using multi-stage deep learning networks and older versions of YOLO model. A Faster R-CNN model was used by \cite{chen2019strawberry} to recognize strawberries in outdoor environment for yield prediction. Similarly, strawberry detection based on YOLOv4-tiny was proposed by \cite{zhang2022real} for outdoor environment. Although it might be difficult to compare the performance of our work with the past studies on open-field strawberry detection because of the use of different datasets and computational platform, our model achieved similar results with faster computational speed, as shown in Table 5. In addition, the reported size of Faster R-CNN (165.7 Mb) is around 9 times bigger than the proposed model. Considering strawberry detection for robotic harvesting, the proposed model focused on detecting strawberries in the three maturity levels of immature, nearly mature, and mature, while the other two studies classified into two maturity levels of mature and immature.  
 
\begin{table}[!ht]
\caption {Performance comparison with open field strawberry detection models }
\center
\begin{tabular}{c c c c c}\hline
method & detected classes & mAP($\%$)$^{a}$ & inference speed per image$^{a}$ & size of models (Mb)\\ \hline
\multirow{3}{*}{\makecell{Faster-RCNN(ResNet 50)\\\citep{chen2019strawberry}}}&flower&\multirow{3}{*}{83.88}&\multirow{3}{*}{113}&\multirow{3}{*}{165.7}\\
&immature&&\\
&mature&&\\
 \hline 
 \multirow{3}{*}{\makecell{Modified YOLOv4-tiny \\\citep{zhang2022real}}}&flower&\multirow{3}{*}{82.25}&\multirow{3}{*}{39.2}&\multirow{3}{*}{no report}\\
&immature&&\\
&strawberry&&\\
 \hline
  \multirow{3}{*}{YOLOv5s-Straw (ours)}&immature&\multirow{3}{*}{80.3}&\multirow{3}{*}{18.1}&\\
&nearly mature&&&18.3\\
&mature&&\\
 \hline
 \multicolumn{4}{l}{$^{a}$mAP refers to mean Average Precision.}\\
\multicolumn{4}{l}{$^{b}$inference speed may vary under different computational environment.}
\end{tabular}\\
\begin{flushleft}

\label{to}
\end{flushleft}
\end{table}
The method proposed in this study focuses on detecting strawberries, but it might not cover all aspects of robotic strawberry harvesting requirements. These include further processing, such as center region detection \citep{he2022detecting}, occlusion classification \citep{he2021detection}, and optimal picking point identification \citep{yu2020real}. To improve robotic harvesting, a gentle picking process is suggested, targeting strawberry stems instead of the whole fruit, which is especially useful in environments where direct fruit detection and picking are not suitable.
The approach differs between open-field and greenhouse/table-top planting due to variations in strawberry canopy architecture and occlusion levels. Detecting strawberries becomes more challenging as they can be fully, partially, or not occluded, leading to varying shapes and sizes. Even leaves can be mistaken as immature fruits, showcasing their complexity. Strawberry color and texture change with maturity, requiring model fine-tuning or a diverse dataset. A dataset with commonly planted varieties could create a robust, widely applicable model. Occlusions like leaves affect later processing steps like picking order or angle. To enhance robotic vision, reconstructing hidden parts of occluded strawberries is suggested. This could improve strawberry detection completeness and enable more accurate picking strategies.

\label{4}
\section{Conclusion and future work}

In this study, we proposed an enhanced version of YOLOv5s called YOLOv5s-Straw for detecting strawberries at three different maturity levels (immature, nearly mature, and mature) in outdoor/field environments. This advancement serves as a crucial stepping stone towards the development of more efficient robotic harvesting systems. The improved network structure incorporated two key modifications. Firstly, the original C3 modules in the backbone networks were replaced with C2f modules, which consist of the Bottleneck module and two Conv modules. This substitution enhanced detection accuracy by facilitating improved gradient information flow. Secondly, the SPPF block in conjunction with CSPNet was optimized to increase the computational efficiency of the output feature map. 
Based on the results, the following specific conclusions are made:

\begin{enumerate}

            \item 
            The proposed model effectively recognized strawberries at varying maturity levels, making it well-suited for robotic harvesting applications in open-field environments The overall mean average precision (mAP) of the proposed model was 80.3\%. Specifically, the average precision (AP) for the immature class was 82. 1\%, whereas the same for nearly mature and mature classes were 73.5\%, and 86.6\%, respectively. 
            
             \item A comparative analysis demonstrated that the mAP of YOLOv5s-Straw surpassed the performance of the other models. Specifically, YOLOv5s-Straw achieved an mAP that was 7.3\%, 2.9\%, 0.9\%, and 1.4\% higher than YOLOv3-tiny, YOLOv5s, YOLOv5s-C2f, and YOLOv8s, respectively. Despise the improved performance, YOLOv5s-Straw maintained a guaranteed detection speed suitable for robotic strawberry harvesting.

        \end{enumerate}

To further enhance the accuracy and effectiveness of strawberry detection in open-field conditions, future work could utilize a diverse strawberry dataset that encompasses various environmental conditions, strawberry varieties, and maturity levels. By training detection models on such a comprehensive dataset, the generalization ability of the models can be improved, leading to better detection performance in real-world scenarios. Additionally, optimizing the detection algorithm can have a significant impact on the accuracy and efficiency of strawberry detection.Furthermore, considering the challenge of strawberry occlusion, the use of multiple cameras from different angles could be investigated for improved visibility. 

\section*{Acknowledgment}
This research was funded by National Science Foundation of the United States (NSF, award\# 1924640) and Washington State University. Any options, findings, and conclusions expressed in this publication are those of the authors and do not reflect any view from NSF and WSU.

\section*{Declaration of Generative AI and AI-assisted technologies in the writing process}
During the preparation of this work the author(s) used Chatgpt in order to check typos and improve readablity of this study. After using this tool/service, the author(s) reviewed and edited the content as needed and take(s) full responsibility for the content of the publication.

%Bibliography
\bibliographystyle{apacite}  
\bibliography{references}

\end{document}